\documentclass[letterpaper, 10 pt, conference]{ieeeconf}

\IEEEoverridecommandlockouts                              
\usepackage[utf8]{inputenc}
\usepackage{graphicx}   
\usepackage{xcolor}
\usepackage{amsmath,amssymb, amsfonts, mathtools}
\mathtoolsset{showonlyrefs}
\usepackage[utf8]{inputenc}
\usepackage[english]{babel}
\usepackage[caption=false,font=footnotesize]{subfig}
\usepackage{mathrsfs}

\usepackage{amsmath,amsthm,amssymb}
\usepackage{mathtools}
\mathtoolsset{showonlyrefs}
\usepackage{mathrsfs}
\usepackage{comment}
\newtheorem{theorem}{Theorem}

\newtheorem{example}{Example}

\DeclareMathOperator*{\argmin}{arg\,min}

\newcommand{\R}{\mathbb{R}}

\newcommand{\tr}{^T}

\newcommand{\mc}{\mathcal}

\definecolor{blue}{rgb}{0.0, 0.0, 1.0}

\DeclareMathOperator*{\minimize}{minimize}
\DeclareMathOperator*{\subjto}{subject\,to}

\title{\LARGE \bf
Adaptive Task Allocation for Heterogeneous Multi-Robot Teams with Evolving and Unknown Robot Capabilities 
}
\author{Yousef Emam$^{1}$, Siddharth Mayya$^{1}$, Gennaro Notomista$^{1}$,  Addison Bohannon$^{2}$, Magnus Egerstedt$^{1}$
	\thanks{*This work was supported by the Army Research Lab through ARL DCIST CRA W911NF-17-2-0181.}
	\thanks{$^{1}$Y. Emam, S. Mayya, G. Notomista and M. Egerstedt are with the Institute for Robotics and Intelligent Machines, Georgia Institute of Technology, Atlanta, GA 30332, USA {\tt\small\{emamy, siddharth.mayya, g.notomista, magnus\}@gatech.edu}}
	\thanks{$^{2}$A. Bohannon is with the CCDC Army Research Laboratory, Aberdeen Proving Ground, MD 21005 USA {\tt\small addison.w.bohannon.civ@mail.mil}}
}
\date{\today}

\begin{document}

\maketitle

\begin{abstract}
For multi-robot teams with heterogeneous capabilities, typical task allocation methods assign tasks to robots based on the suitability of the robots to perform certain tasks as well as the requirements of the task itself. However, in real-world deployments of robot teams, the suitability of a robot might be unknown prior to deployment, or might vary  due to changing environmental conditions. This paper presents an adaptive task allocation and task execution framework which allows individual robots to prioritize among tasks while explicitly taking into account their efficacy at performing the tasks---the parameters of which might be unknown before deployment and/or might vary over time. Such a \emph{specialization} parameter---encoding the effectiveness of a given robot towards a task---is updated on-the-fly, allowing our algorithm to reassign tasks among robots with the aim of executing them. The developed framework requires no explicit model of the changing environment or of the unknown robot capabilities---it only takes into account the progress made by the robots at completing the tasks. Simulations and experiments demonstrate the efficacy of the proposed approach during variations in environmental conditions and when robot capabilities are unknown before deployment.

\end{abstract}

\section{Introduction}
\label{sec:intro}

Consider a disaster relief scenario where a heterogeneous team of robots (e.g., drones, autonomous ground vehicles, underwater robots, etc.) is deployed to search and rescue victims. The robots, each with their own specialized capabilities, must accomplish multiple tasks, such as reconnaissance, coverage, and navigating to certain locations with the aim of rescuing victims. In situations such as the example mentioned above, there exists a need to assign tasks to robots so as to satisfy certain task requirements while explicitly taking into account the specialized capabilities of the individual robots \cite{parker1994heterogeneous,taxonomy}.\par
The multi-robot task allocation (MRTA) literature (e.g., see \cite{taxonomy,khamis2015multi} and references within) has addressed the task allocation problem for heterogeneous teams using a variety of different approaches including stochastic allocation \cite{berman2009optimized}, market based approaches \cite{dias2006market} or cost-based assignment \cite{balinski1985signature, zhang2012centralized}. In \cite{notomista2019optimal}, the authors present a task allocation and execution framework which allows task assignments to explicitly account for the energy requirements of the tasks as well as global specifications on the task allocation. The heterogeneous capability of each robot towards each task is captured by a \emph{specialization} parameter which encodes the effectiveness of the robot at performing the task. This is included within an optimization problem which is solved at each time step. \par 

However, in certain deployment scenarios, the specialization of a robot towards a particular task might be unknown prior to deployment of the robot team, or might vary due to changes in the environmental conditions. For example, the effectiveness of a robot at tracking a target might depend on the speed and altitude of the target---parameters which might be unknown apriori. Alternatively, wheeled robots might be rendered inoperative in an outdoor deployment scenario if, for instance, sudden rain causes certain terrain to become impassable. Not only can such factors affect the ability of robots to perform certain tasks, explicitly modeling or predicting such effects might be impossible. In such scenarios, non-adaptive task allocation schemes can generate deficient and sometimes infeasible assignments given the real-time conditions. \par 
Motivated by such scenarios, and leveraging the task-allocation formulation presented in \cite{notomista2019optimal}, this paper develops an adaptive task allocation scheme which dynamically updates the specialization parameters of the robots based on their current performance at accomplishing the tasks. By doing so, we allow the task allocation framework to dynamically re-assign tasks to robots, thereby compensating for the lack of apriori knowledge on robot capabilities and mitigating the negative effects of environmental variations---with the aim of allowing the multi-robot team to make progress towards accomplishing the tasks. \par 

In this paper, we evaluate the current capabilities of the robots based on their efficacy at performing the tasks, thus removing the need for explicitly modeling environmental variations or predicting the performance of robots. More specifically, we leverage the fact that, in the framework presented in \cite{notomista2019optimal}, task execution is encoded through the minimization of a cost function. This allows us to measure the deviation between the robots expected progress towards the task accomplishment and their actual progress. The specialization parameters can be updated for each robot according to this observed deviation. Tasks can then be dynamically re-assigned with the updated specialization parameters. \par

\section{Literature Review} \label{sec:lit_rev}

Adaptive task allocation for heterogeneous multi-agent systems is a well-studied topic \cite{parker1994heterogeneous, lerman2006analysis,iocchi2003distributed, palmieri2018self, FatimaS2001Atra, iijima2017adaptive, 4803959}. A common trend in these existing frameworks is to let each agent have a preference towards the various tasks depending on its current state and corresponding objective (e.g. \cite{iocchi2003distributed, iijima2017adaptive}). The corresponding utility function of each agent is usually pre-determined based on its role or pre-defined strategy and serves to represent the effectiveness of the agent at performing a task in real-time. The task allocation is then performed using market-based, centralized, decentralized, or learning protocols based on the values of the robots' so-called utility functions in real-time. For example, in \cite{iijima2017adaptive}, the multi-agent system is composed of computational entities each having specific functionalities and resources. The authors formulate a method based on $\varepsilon$-greedy Q-learning to assign the incoming stream of tasks to agents depending on their respective preferences. \par 
In contrast to the above mentioned approaches, this paper utilizes a task allocation framework that simultaneously accounts for the assignment of tasks to robots as well as the execution of the tasks itself as first presented in \cite{notomista2019optimal}. In our approach, task execution is encoded as the minimization of a cost function, which leads to a natural interpretation of the effectiveness of robots at performing tasks. Furthermore, our framework assigns tasks to robots taking into account the energy required to perform the tasks, the global specifications of the task allocation (e.g. the desired distribution of robots over the tasks), and the real-time effectiveness of the robots at performing the tasks. It is important to note that this approach requires a high-level specification of the desired assignment which can be provided by any algorithm aimed at solving any instance of the Instantaneous Assignment (IA) problems (e.g. \cite{coalitions,4028037,vig2006market}) as defined in \cite{taxonomy}. This is possible because the task allocation component in \cite{notomista2019optimal} can be viewed as a variation of the Multi-Robot tasks Multi-Task robots Instantaneous Assignment (MR-MT-IA) problem. \par

\section{Multi-Task Execution and Prioritization}
\label{sec:background}
This section briefly presents a modified version of the task execution and prioritization framework first introduced in \cite{notomista2018constraint} and \cite{notomista2019optimal}, which will be used in the next section to develop the proposed adaptive task allocation framework. We begin by introducing a task execution framework where each robot minimizes its control effort subject to a constraint which enforces the execution of the task. Then, we demonstrate that this formulation also allows the robots to prioritize among the various tasks. \par
\subsection{Task Execution}
Consider a multi-robot team consisting of $N$ robots which are to execute $M$ tasks, denoted by $T_1,\ldots,T_M$. For the sake of generality, we assume that each robot $i$, where $i\in \{1,\ldots,N\} \overset{\Delta}{=} \mc N$, in the multi-robot system can be modeled as the following control-affine dynamical system 
\begin{equation}
    \label{eq:controlAffine}
    \dot x_i = f(x_i) + g(x_i) u_i,
\end{equation}
where $x_i \in X \subseteq \mathbb{R}^n$ is the state, $u_i \in U \subseteq \mathbb{R}^m$ is the input, and $f$ and $g$ are locally Lipschitz continuous vector fields. Following the formulation in \cite{notomista2018constraint}, this paper considers a specific class of tasks which can each be encoded via a positive definite, continuously differentiable cost function. The execution of a single task is then identified as the minimization of its corresponding cost function. Such a formulation of multi-robot tasks can be used to encode a wide variety of tasks such as formation control, environmental surveillance, and path following \cite{cortes2017coordinated}.  Let $V_{ij} \colon \mathbb{R}^n \rightarrow \mathbb{R}$ denote the component of the cost function corresponding to task $T_j$ associated with robot $i$ \cite{cortes2017coordinated}.     Mathematically, the execution of task $T_j$ by robot $i$ can be encoded as the following minimization problem
\begin{align}
\label{eq:minJ}
    \minimize_{u_i} \quad & V_{ij}(x_i) \\
    \subjto \quad &\dot{x_i} = f(x_i) + g(x_i) u_i.
\end{align}
In \cite{notomista2018constraint}, an alternative formulation for the execution of tasks was developed, which encoded the tasks in the constraint as opposed to the cost. More specifically, it was shown that generating a control signal $u_i(t)$ by solving \eqref{eq:minJ} was mathematically equivalent to solving the following constraint-based optimization problem
\begin{equation}
\label{eqn:const_opt}
\begin{aligned} 
\minimize_{u_i,\delta_{ij}}~~&\|u_i\|^2 + |\delta_{ij}|^2 \\
\subjto~~&L_f h_{ij}(x_i) + L_g h_{ij}(x_i) u_i \geq -\gamma(h_{ij}(x_i)) - \delta_{ij},
\end{aligned}
\end{equation}
where $h_{ij}(x_i) = -V_{ij}(x_i)$ is a \emph{Control Barrier Function (CBF)} \cite{cbftutorial}, $\delta_{ij}$ is a slack variable which represents the extent to which the constraint corresponding to the task execution of task $T_j$ can be violated, $L_{f}h_{ij}(x_i)$ and $L_{g}h_{ij}(x_i)$ denote the Lie derivatives of $h_{ij}$ in the directions $f$ and $g$ respectively, and $\gamma$ is an extended class-$\mathcal{K}$ function \cite{cbftutorial}. In \eqref{eqn:const_opt}, the execution of the task is enforced by driving the system towards a ``safe-set" which encodes the completion of the task (i.e., minimization of the corresponding cost function $V_{ij}$). This constraint-based formulation is shown to be advantageous compared to \eqref{eq:minJ} in terms of the long-term operations of robots in dynamic environments \cite{egerstedt2018robot}. \par

Next, we discuss the heterogeneous task allocation component of the framework, with the ultimate aim of developing the adaptive task allocation mechanism in Section~\ref{sec:adaptiveTA}.

\subsection{Task Allocation for Heterogeneous Teams}
\label{subsec:optimalTA}
In this section, we introduce constraints on the slack variables $\delta_{ij}$ to allow the robots to prioritize some tasks over others. For example, the specification that a given task $T_{j}$ is to be executed with the highest priority by robot $i$ can be encoded by the following constraint:
\begin{equation} \label{eq:deltanm}
    \delta_{ij} \leq \delta_{ik} \quad \forall k \in \mathcal{M},
\end{equation}
where $\delta_{ik}$ denotes the extent to which robot $i$ can violate the constraint corresponding to task $T_k$ and $\mc M = \{1,\ldots,M\}$. Such constraints can be written in a more general form as $\mc P\delta_{i} \leq 0$, where $\delta_i = [\delta_{i1},\ldots,\delta_{iM}]$, $\mc P \in \mathbb{R}^{q \times m}$ is called the prioritization matrix, and $q$ denotes the number of desired constraints.\par 
The task allocation framework presented above aims at allocating tasks among robots according to some desired specifications while also taking into account the energy consumed by the robots as well as the heterogeneity in their capabilities. Next, we discuss this formalism which is required to introduce our adaptive task allocation approach.
\subsubsection{Heterogeneity} To account for the heterogeneity of the robot team, the task allocation framework in \cite{notomista2019optimal} factors in the suitability of each agent to the various tasks. To this end, let $s_{im}\ge 0$ denote a \textit{specialization} parameter corresponding to the suitability of robot $i$ for executing task $T_m$. In other words, $s_{im}>s_{in}$ implies that robot $i$'s capabilities are better suited to execute task $T_m$ than task $T_n$.  This parameter will play a core role in Section~\ref{sec:adaptiveTA}, where it will be updated based on the current performance of the robots towards the various tasks. The \textit{specialization} matrix $S_i$ is then defined as follows $S_i = diag([s_{i1},\ldots,s_{iM}])$.
Since $S_i$ is a diagonal matrix whose entries are all non-negative, we can define the seminorm $\|\cdot\|_{S_i}$ by setting $\|x_i\|_{S_i}^2 = x_i\tr S_i x_i,~x_i\in\R^n$. This corresponds to weighting each component of $x_i$ by the corresponding entry of $S_i$. 

\subsubsection{Global Specifications}
To increase the flexibility of the algorithm, we would like to be able to specify a desired robot allocation in terms of the fractions of robots that should be allocated to each task. To this end, let $\pi_m^*$ denote the desired fraction of robots that need to perform task $T_m$ with highest priority. Then, we refer to $\pi^* = [\pi_1^*,\pi_2^*,\ldots,\pi_M^*]^T$ as the global task specification for the team of robots. \par

In order to achieve such a desired specification, we introduce the variable $\alpha_i = [\alpha_{i1},\ldots,\alpha_{iM}]^T \in \{0,1\}^M$, whose entries indicate the priorities of the tasks for robot $i$ (i.e. $\alpha_{im} = 1$ $\Longleftrightarrow$ task $T_m$ has the highest priority for robot $i$).
At any given point in time, only one element of $\alpha_i$ can be non-zero. 
Moreover, given the priority constraints in \eqref{eq:deltanm} and the definition of $\alpha_{im}$, we would like the following implication to hold,
\begin{equation}\label{eq:alphadelta}
\alpha_{im} = 1 \quad\Rightarrow\quad \delta_{im} \leq \frac{1}{\kappa}\delta_{in}\quad \forall n\in \mathcal{M},~n\neq m,
\end{equation}
where $\kappa > 1$ allows us to encode how the task priorities impact the relative effectiveness with which robots perform different tasks. 
Let $\alpha = [\alpha_1\tr,\alpha_2\tr,\dots,\alpha_N\tr]\tr \in \{0,1\}^{NM}$ represent the vector containing the task priorities for the entire multi-robot system. Then, the task prioritization of the team at any given point in time can be compactly written as:
\begin{equation}
\label{eq:pi_h}
\pi_h(\alpha) = \frac{1}{N} \begin{bmatrix}
P_1, P_2, \ldots, P_N
\end{bmatrix} \alpha,
\end{equation}
where $P_i = S_i S_i^\dagger$ and $S_i^\dagger$ is the Moore-Penrose inverse of $S_i$.
The expression in $\pi_h(\alpha)$ in \eqref{eq:pi_h} implies that the contribution of robots prioritizing tasks for which they are not suitable is discounted in calculating the current allocation of the robot team \cite{notomista2019optimal}.
Considering all the factors presented above yields the following optimization program:
\begin{subequations} \label{eq:allocationalgorithm}
\begin{align}
\minimize_{u,\delta,\alpha} ~& C\|\pi^* - \pi_h(\alpha)\|_{T}^2 + \sum_{i = 1}^{N} \Big( \|u_i\|^2 + l \|\delta_i \|_{S_i}^2 \Big) \label{eq:miqp:a}\\ 
\subjto~~&L_f h_{m}(x) + L_g h_{m}(x) u_i \geq -\gamma(h_{m}(x)) - \delta_{im} \label{eq:miqp:b} \\
&\mc P\delta_i \le \Omega(\alpha_i,\kappa,\delta_{max}) \label{eq:miqp:c}\\
&\boldsymbol{1}\tr\alpha_i = 1 \label{eq:miqp:d}\\ 
&\|\delta_i\|_\infty \leq \delta_{max} \label{eq:miqp:e}\\
&\alpha \in \{0,1\}^{NM} \label{eq:miqp:f}\\
&\hspace{3cm}\forall i \in \mc N,~\forall n,m \in \mc M.
\end{align}
\noeqref{eq:miqp:a}\noeqref{eq:miqp:b}\noeqref{eq:miqp:c}\noeqref{eq:miqp:d}\noeqref{eq:miqp:e}\noeqref{eq:miqp:f}
\end{subequations}
\hspace{-0.2cm}where $x_i$ and $u_i$ denote the state and input of robot $i$ respectively, and $x = [ x_1 \tr x_2 \tr \ldots x_N\tr]$ denotes the ensemble state.  In \eqref{eq:miqp:a}, $C$ and $l$ are scaling constants allowing for a trade-off between meeting the global specifications and allowing individual robots to expend the least amount of energy possible. The term $\|\delta_i \|_{S_i}^2$ accounts for the heterogeneity of the robot team by not penalizing robots' slack variables corresponding to tasks for which they are not suitable (i.e. $s_{ij} = 0$). Using the matrix $\Omega$, the constraint \eqref{eq:miqp:c} encodes the relation described in \eqref{eq:alphadelta} (\cite{notomista2019optimal}).

Finally, to measure the difference between the desired and actual task prioritization, we introduce the seminorm defined by $\|\cdot\|_{T}$, where $T\ge0$ is a diagonal matrix denoting the relative importance of meeting the required distribution for each task respectively. We now show an example of the task prioritization framework explained above. Following this, the next section presents the proposed adaptive task allocation framework.

\begin{example} \label{exmp:sim0}
\begin{figure}[t!]
    \centering
    \subfloat[]{\label{subfig:sim0_init}\includegraphics[width=0.24\textwidth]{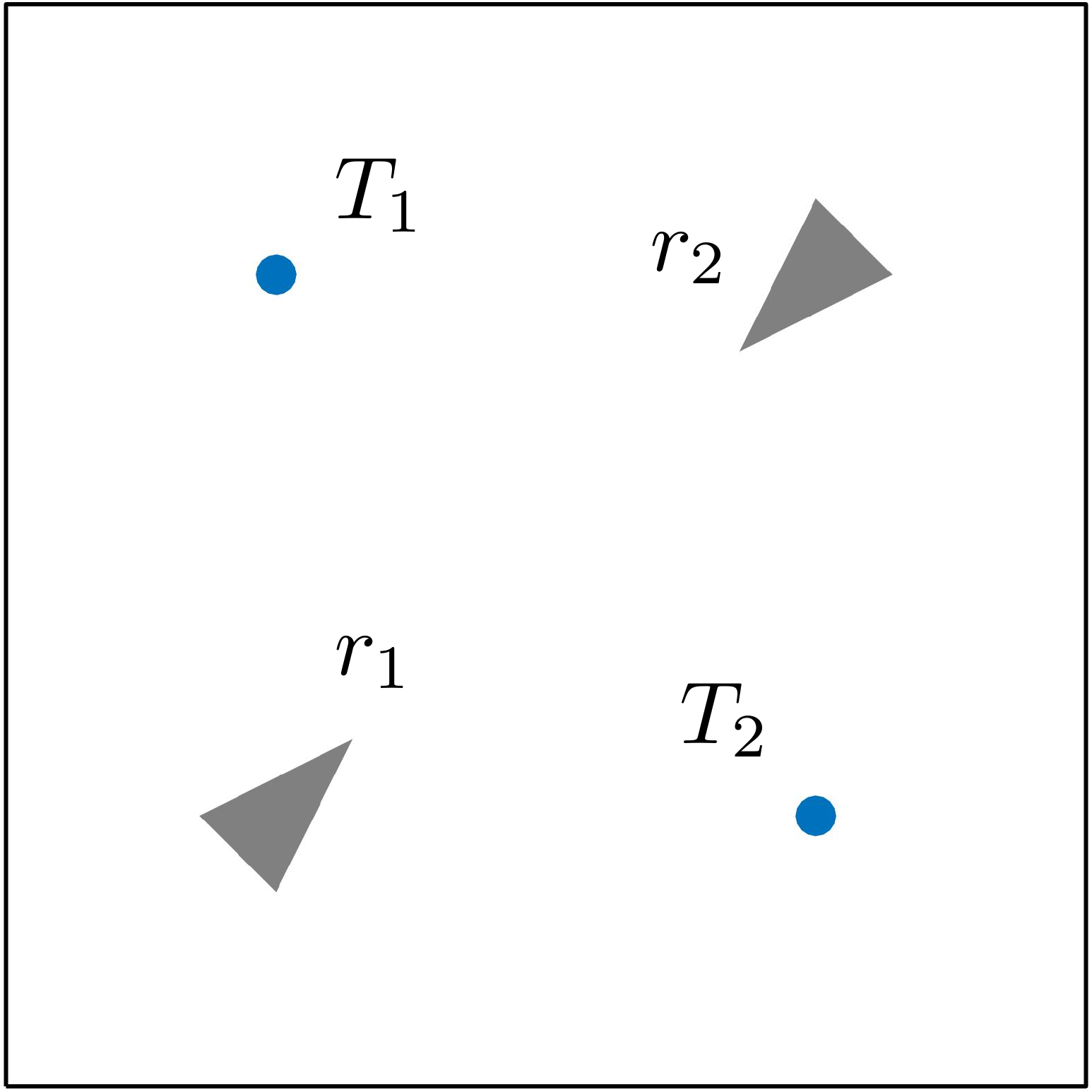}}\hfill
    \subfloat[]{\label{subfig:sim0_traj}\includegraphics[width=0.24\textwidth]{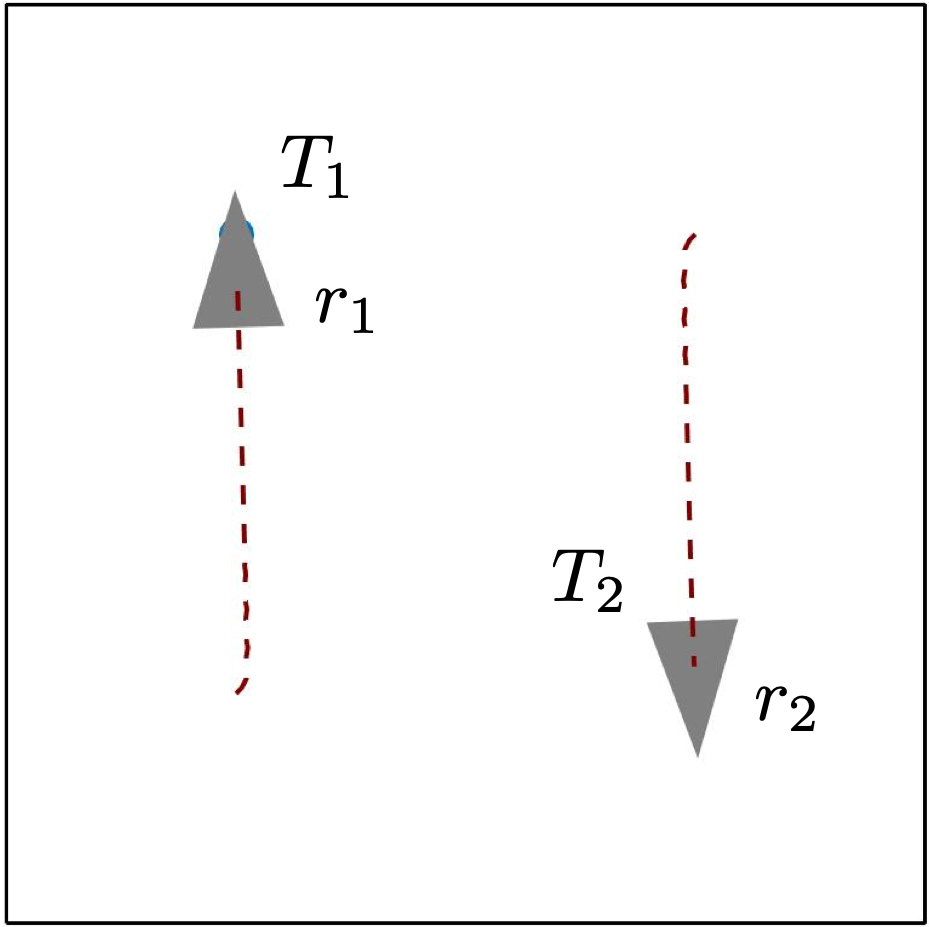}}\\
    \caption{Simulation corresponding to the problem stated in Example~\ref{exmp:sim0} where $2$ robots are to execute $2$ go-to-goal tasks. It is stated that although $r_2$ can accomplish either tasks, $r_1$ can only execute $T_1$. The robots' trajectories are shown on the right by the dashed-red lines. Indeed, the framework in \eqref{eq:allocationalgorithm} correctly assigns the robots so that both tasks are successfully executed.}
    \label{fig:sim0}
\end{figure}
Consider the scenario in which $2$ robots, $r_1$ and $r_2$ are to be assigned to $2$ tasks, $T_1$ and $T_2$, each requiring a single robot to drive to a goal location (see Fig.~\ref{fig:sim0}). The two robots are heterogeneous insofar as $r_1$ can only accomplish $T_1$, whereas $r_2$ can accomplish both $T_1$ and $T_2$. The corresponding specialization matrices for $r_1$ and $r_2$ would be then defined as $S_1~=~diag([1, 0])$ and $S_2~=~diag([1, 1])$. 
The global specification $\pi^*~=~[1/2, 1/2]^T$ encodes the fact that only one robot is required to execute each of the tasks. According to \eqref{eq:pi_h}, the allocation of $T_1$ to $r_1$ (i.e. $\alpha_{11} = 1$) and $T_2$ to $r_2$ (i.e. $\alpha_{22}~=~1$), would yield $\pi_h(\alpha)~=~[1/2, 1/2]^T$. On the other hand, the prioritization of $T_1$ by $r_2$ and $T_2$ by $r_1$ would yield $\pi_h(\alpha)~=~[1/2, 0]^T$ due to the inability of $r_1$ to execute $T_2$ as reflected in $S_1$. Therefore, as can be seen in Fig.~\ref{fig:sim0}, solving the optimization algorithm \eqref{eq:allocationalgorithm}, results in the former allocation, which satisfies the global specification encoded in $\pi^*$.
\end{example}




\section{Adaptive Specialization for Dynamic Environments}
\label{sec:adaptiveTA}

As discussed in Section~\ref{sec:intro}, this paper focuses on the scenario where the capabilities of the robots are unknown or might be affected by varying environmental conditions. This implies that the specialization parameters $s_{ij}$ which encode the effectiveness of robot $i$ at performing task $T_j$ might be unknown or might vary over time---which can severely compromise the effectivness of the task allocation and execution. To this end, Section~\ref{subsec:GenAdaptiveTA} proposes an update law which modifies the specialization parameters $s_{ij}$ as a function of the progress of the robots towards the completion of the tasks.\par

\subsection{Adaptive Update Law}
\label{subsec:GenAdaptiveTA}
Given that the task allocation problem is solved at fixed intervals $dt$, we update the continuous time system at $(kdt)_{k\in\mathbb N}$ and we index the discrete time system with $k$.
We would like to update the parameters $s_{ij}$ at each time step $k$ based on the difference between the expected and actual effectiveness of the task allocation and execution framework. We assume that this difference manifests itself in terms of variations in the dynamical model of the robot. For example, a wheeled robot traversing muddy terrain will cover less distance than predicted according to the robots nominal dynamics. To this end, we introduce notation to differentiate between the modeled/expected state of a robot and its actual state. At time $k$, let $x_i^{\text{act}}[k]$ denote the actual state of robot $i$, and $x_i^{\text{sim}}[k]$ denote the simulated state which assumes that the robot obeyed its nominal dynamics. The simulated states are updated as follows:
\begin{align}
    \label{eq:x_sim}
    x_i^{\text{sim}}[k] &= x_i^{\text{act}}[k-1] \\  &+  (f(x_i^{\text{act}}[k-1]) + g(x_i^{\text{act}}[k-1])u_i^{\ast}[k-1])dt.
\end{align}\par 
 The formulation of tasks used to derive \eqref{eq:allocationalgorithm} naturally lends itself to measuring the difference between the simulated and the actual progress towards the completion of task $T_j$ by robot $i$ at time step $k$ as follows:
\begin{equation}
    \label{deltaV}
    \Delta V_{ij}[k] = V _{ij}(x_i^{\text{sim}}[k]) - V_{ij}(x_i^{\text{act}}[k]),
\end{equation}
where $V_{ij}(x_i^{\text{sim}}[k])$ and $V_{ij}(x_i^{\text{act}}[k])$ are the simulated and actual cost function values of agent $i$ for task $j$ at time step $k$. It is important to note that we assume that each cost function $V_j$ is decomposable into the respective contributions $V_{ij}$ of each robot $i$ which can be computed locally. This assumption is task dependent, and holds for a large number of coordinated control tasks (e.g. consensus, formation control, coverage \cite{cortes2017coordinated}). As such, if $V^{\text{act}}_{ij}[k] > V^{\text{sim}} _{ij}[k]$, robot $i$'s actual effectiveness at accomplishing task $j$ is lower than anticipated. Consequently, one can model the specialization of robot $i$ at task $j$ to evolve according to the following update law:
\begin{equation}
    \label{eq:spUpdate1}
    s_{ij}[k + 1]= s_{ij}[k] + \beta_{1} \alpha_{ij}[k]  \Delta V_{ij}[k],
\end{equation}
where $\beta_{1} \in \R_{>0}$ is a constant controlling the update rate. This update law allows us to account for the dynamical variations in the environmental conditions by updating the specialization matrix on the fly according to the agents' performance. Note that the update only occurs for tasks to which the robots are assigned since $\alpha_{ij}[k] = 1$ if and only if robot $i$ is assigned to task $j$ at time step $k$.

\begin{example}
\label{exmp:sim1}
\begin{figure}
    \centering
    \subfloat[]{\label{subfig:sim1_traj}\includegraphics[width=0.24\textwidth]{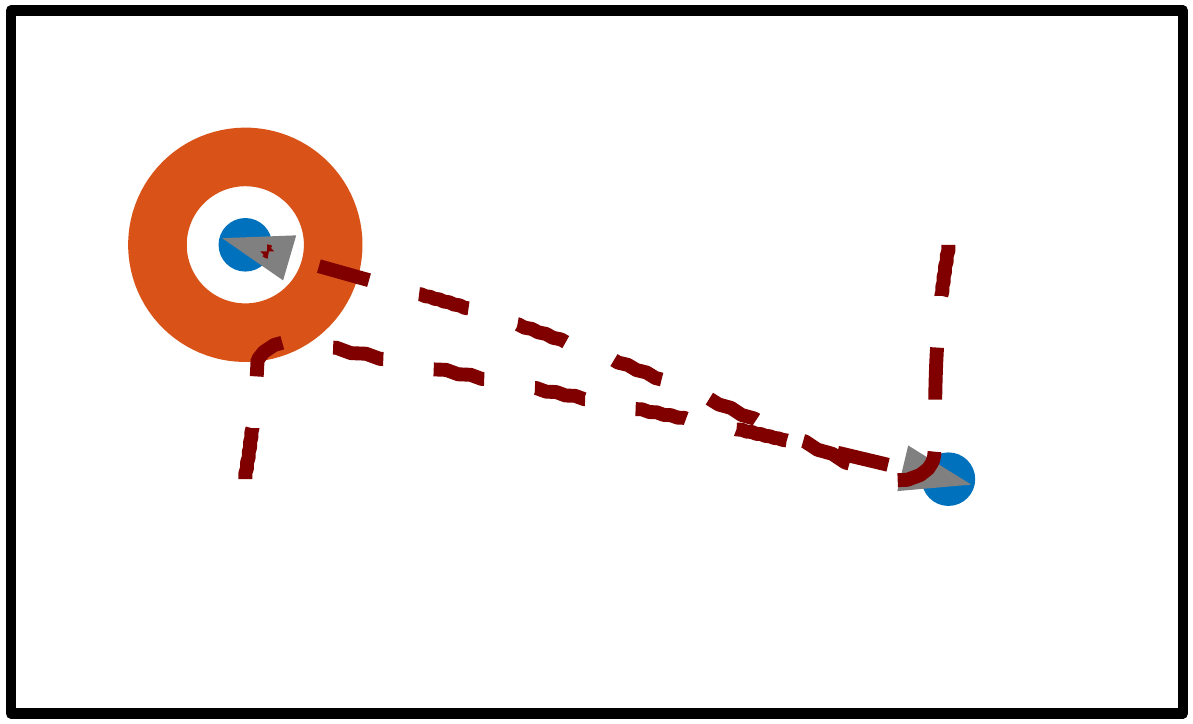}}
    \subfloat[]{\label{subfig:sim1_S}\includegraphics[width=0.24\textwidth]{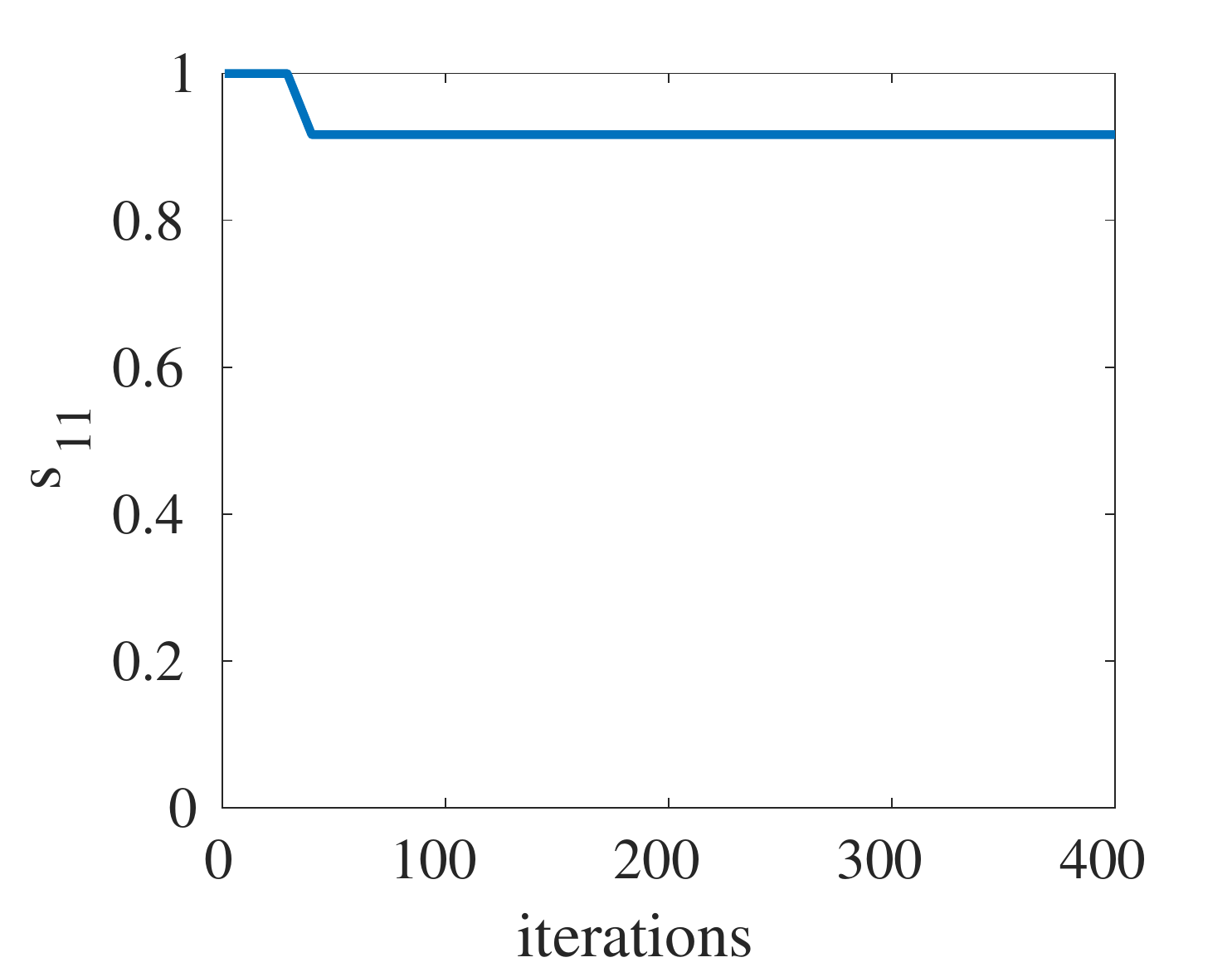}}
    \caption{Simulated experiment showing the application of the adaptive update law \eqref{eq:spUpdate1}. Two robots (gray triangles) are tasked with reaching two goal locations (blue dots). To simulate heterogeneity and unknown environmental conditions, we impose the constraint that only robot $2$ can traverse the red-region. The latter constraint is not reflected in the specialization matrix at initialization. However, thanks to the update law \eqref{eq:spUpdate1}, the tasks are reallocated due to the change in the robot specializations as illustrated by the change in trajectory of the robots (dashed lines).}
    \label{fig:sim1}
\end{figure}
To illustrate the use of the proposed update law in \eqref{eq:spUpdate1}, consider the example depicted in Fig.~\ref{fig:sim1}. Here, two robots (gray triangles) are tasked with visiting two goals (blue dots). The red ring-shaped zone prevents robot 1 from reaching goal 1, while robot 2 is able to overcome it. This could represent, for instance, a region where the ground offers a much lower friction, so that robots not equipped for this type of terrain are not able to drive through. The presence of the red region is not captured by the model, therefore the task allocation algorithm initially assigns task 1 to robot 1 and task 2 to robot 2. When robot 1 gets blocked in the red region while trying to reach goal 1, the value of $\Delta V_{11}[k]<0$. Consequently, the value of $s_{11}[k+1]$ is smaller than $s_{11}[k]$ and it decreases up to the point at which solving \eqref{eq:allocationalgorithm} results in a switch in the task allocation. Task 2 is now assigned to robot 1 and task 1 is assigned to robot 2 due to the fact that robot 1 is not suitable to execute task 1 as expected anymore.

The trajectories of the robots over the course of the simulated experiments are depicted as red dashed lines in Fig.~\ref{subfig:sim1_traj}. Figure~\ref{subfig:sim1_S} shows the graph of the component of the specialization matrix $S_1$ related to task 1, which reflects the behavior discussed above.
\end{example}

\begin{figure}[t!]
    \centering
    \includegraphics[width=0.5\textwidth]{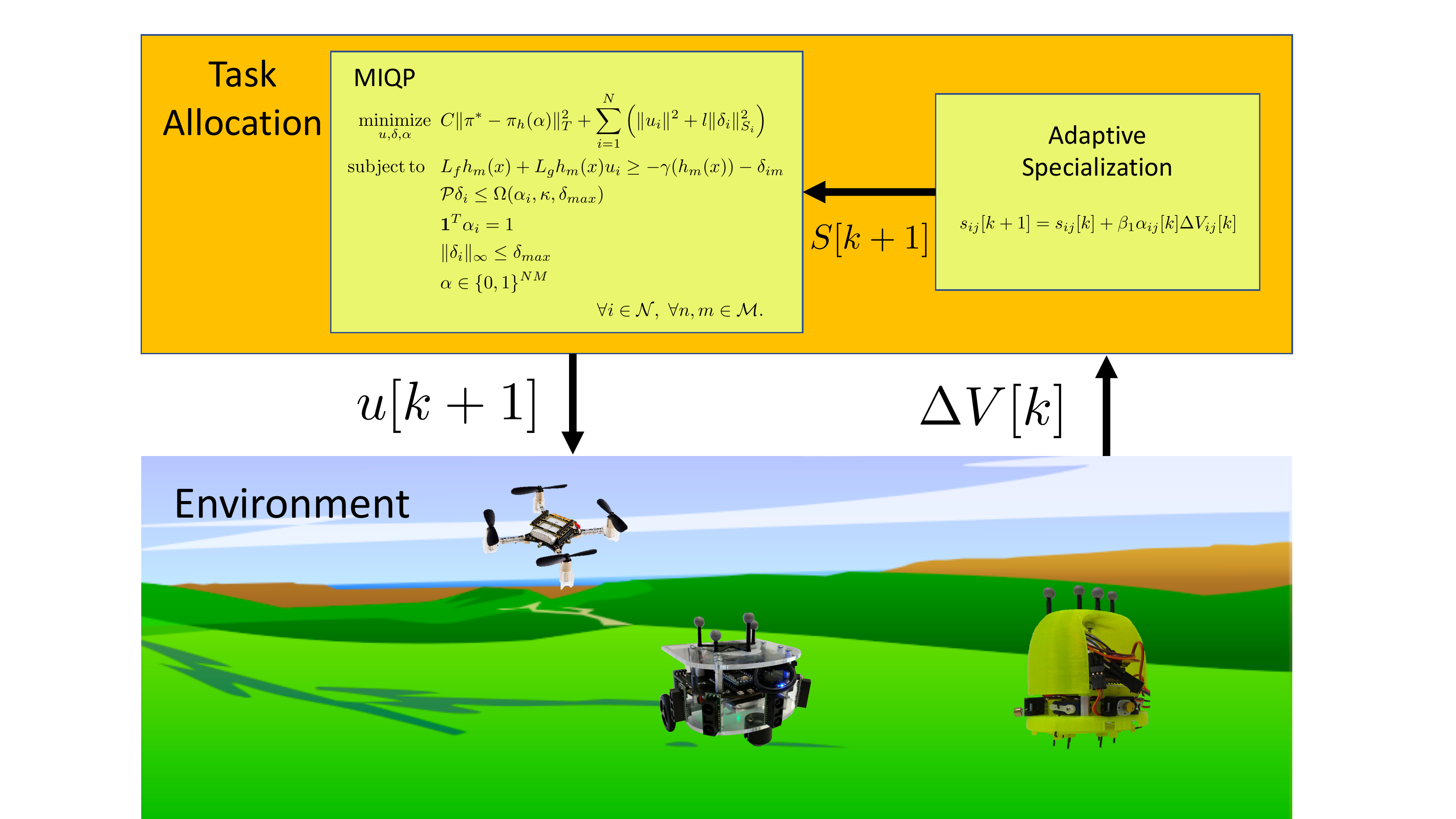}
    \caption{A figure illustrating the proposed feedback loop between the task allocation framework \eqref{eq:miqp:a}, the environment, and the adaptive specialization update. The effectiveness of the robots at each task is measured as $\Delta V[k]$, which is passed to the adaptive specialization update law. Once the new specialization parameters are computed, the task allocation MIQP is solved and the inputs are sent to the robots.}
    \label{fig:feedbackLoop}
\end{figure}

Figure~\ref{fig:feedbackLoop} pictorially shows how this update law for $S_i$ results in a feedback loop between the robots, the changing environment and the task allocation algorithm.

In this subsection, we formulated an update law which mitigates the effects of environmental changes by adapting the specialization matrix on the fly. Specifically, we make no assumptions on the nature (e.g. temporal or spatial) of such variations. In situations where the specializations of the robots are well-known a priori with respect to some nominal conditions, and the environmental changes are expected to be solely temporary or spatially confined, it might be desirable to restore the values of the specialization parameters $s_{ij}$ to the nominal ones which we will refer to as $\bar{s}_{ij}$. Next, we discuss a modification to the update law to incorporate this scenario.

\subsection{Special Case: Isolated Environmental Variations}
\label{subsec:SCadaptiveTA}

In this subsection, we address the case where environmental changes are limited in time or space and the capabilities of the robots are known a priori. Specifically, let $\bar{s}_{ij}$ be the true specialization parameter of robot $i$ executing task $j$ given by a high-level planner (e.g. a method from \cite{taxonomy}) which takes into account nominal environmental conditions. Environmental disturbances are considered spatially and temporally limited when the following condition is satisfied:
\begin{equation}
    \frac{1}{|U|}\int_{\mc D\times U \times \mc T} \mathbf{1}_{\{|\dot{x}(t)-f(x(t))-g(x(t))u|>\varepsilon\}} \ll |\mc T|\cdot|\mc D|
\end{equation}
where $\mathbf{1}$ is the indicator function, $\varepsilon \in \mathbb{R}_{\geq0}$ is a threshold above which we define an environmental variation to have occurred, $\mc T$ and $\mc D$ are the time interval and the domain, respectively, over which the robots are deployed. In this case, it is desirable for the specialization parameters to track the nominal parameters given by $\bar{s}_{ij}$. This can be accomplished through adding a correction term to the update law, yielding

\begin{align}
    \label{eq:spUpdate2}
    s_{ij}[k + 1] = s_{ij}[k] + \beta_1 \alpha_{ij} \Delta V_{ij}[k]  \\ 
    + \beta_2 \sum_{l=0}^{k} (\bar{s}_{ij}  - s_{ij}[l]) dt,
\end{align}
where $\beta_2 \in \mathbb{R}_{>0}$ is the capability recuperation rate. It is important to note that $\beta_1 >> \beta_2$ so as to ensure that the recuperation occurs at a much slower rate that the live update, ensuring that the effects of both terms do not cancel out. 

\begin{example}
\begin{figure}
    \centering
    \subfloat[]{\label{subfig:sim2_traj}\includegraphics[width=0.24\textwidth]{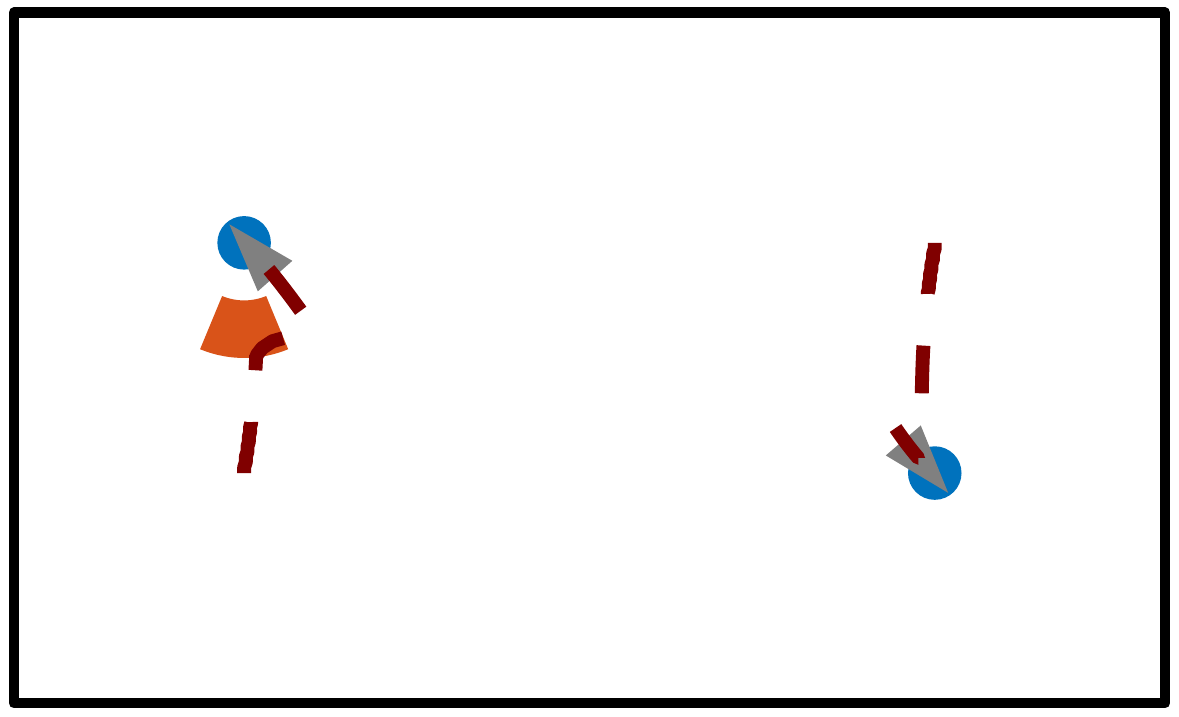}}
    \subfloat[]{\label{subfig:sim2_S}\includegraphics[width=0.24\textwidth]{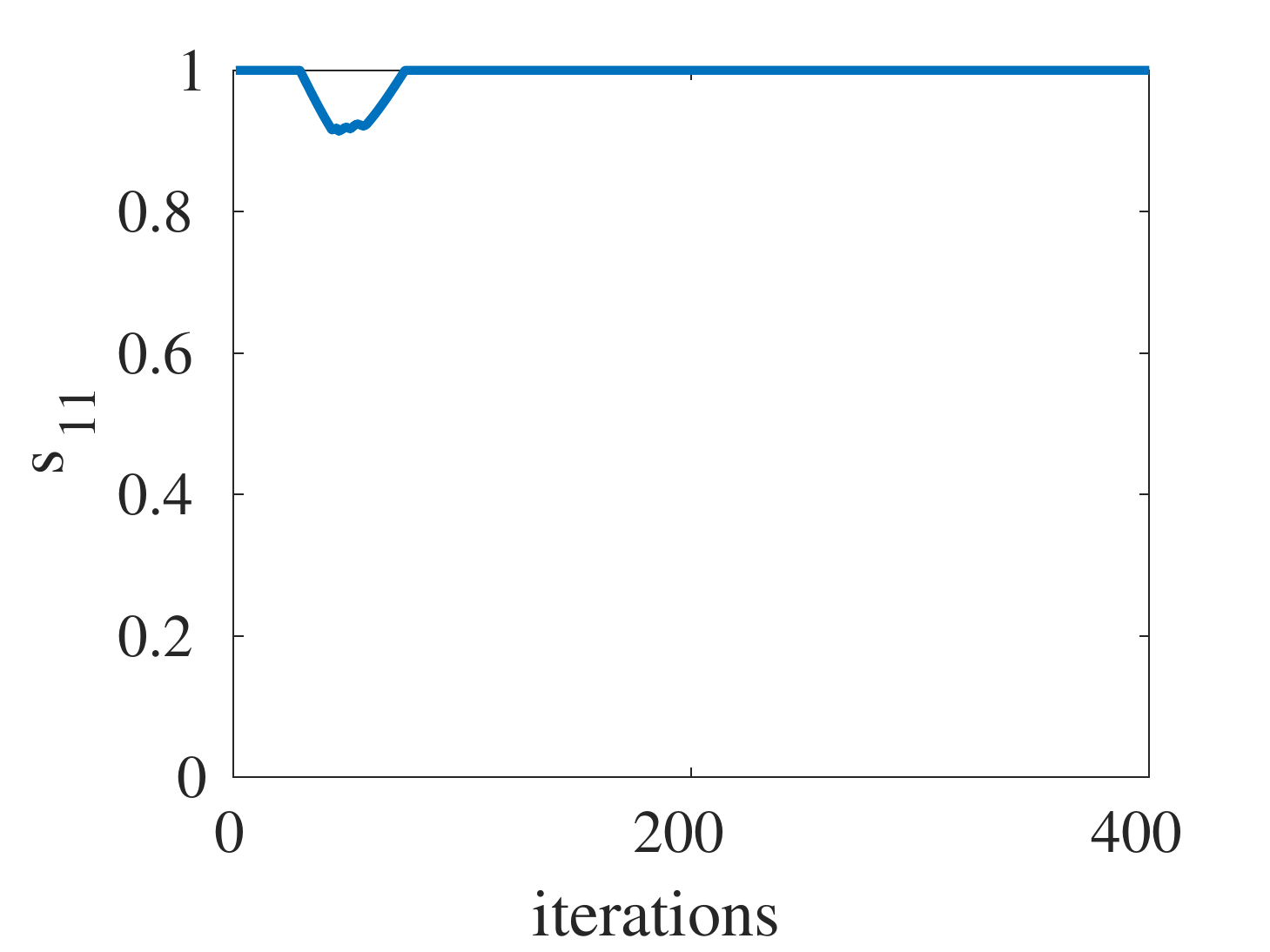}}
    \caption{Variation of the simulated experiment shown in Fig.~\ref{fig:sim1}. Unlike that case, here robot 1 is able to reconsider its initial specialization matrix thanks to the new update law given in \eqref{eq:spUpdate2}. This capability is particularly suitable for scenarios where the environmental variations are confined either in space or in time. These situations are represented in this case by a smaller red region which prevents robot 1 from reaching goal 1 only in case it comes from a specific direction. The robots' trajectories resulting by the execution of the optimal task allocation algorithm \eqref{eq:allocationalgorithm} are depicted as dashed lines in Fig.~\protect\ref{subfig:sim2_traj}.}
    \label{fig:sim2}
\end{figure}
A modification of the scenario presented in Example~\ref{exmp:sim1} is reported in Fig.~\ref{fig:sim2}. In this case, the region that prevents robot 1 from reaching goal 1 is reduced in space and only obstructs the robot if it comes from a specific direction. Initially, robots 1 and 2 are assigned to tasks 1 and 2, respectively. Similarly to what happens in Example~\ref{exmp:sim1}, as robot 1 is blocked by the red region, the entry of its specialization matrix corresponding to task 1 decreases up to the point when the task allocation is switched, i.\,e. robot 1 is assigned to task 2 while robot 2 executes task 1. However, due to the effect of the term $\beta_2 \sum_{l=0}^{k} (\bar{s}_{ij}  - s_{ij}[l]) dt$ in \eqref{eq:spUpdate2}, the initial vale of $s_{11}$ is restored and, consequently, robot 1 gets reassigned to task 1 by the task allocation algorithm \eqref{eq:allocationalgorithm}. This behavior is desirable in the presented scenario since the environment variations are limited in space, thus allowing the robots to perform both tasks without the need of switching allocation as in Example~\ref{exmp:sim1}.

Figure~\ref{subfig:sim2_S} shows the value of $s_{11}$ over the course of the simulated experiment. The initial decrease is due to the proportional term $\beta_1 \alpha_{ij} \Delta V_{ij}[k]$ in \eqref{eq:spUpdate1} whose effect is then counteracted by the integral term $\beta_2 \sum_{l=0}^{k} (\bar{s}_{ij}  - s_{ij}[l]) dt$. Figure~\ref{subfig:sim2_traj} illustrates the trajectory of the robots obtained by executing the optimal control input solution of \eqref{eq:allocationalgorithm}. 
\end{example}

In case a permanent or major change in the capability of a robot does occur, a high-level planner (e.g. a method from \cite{taxonomy}) can generate new nominal specialization parameters $\bar{s}_{ij}$ which can be tracked using the update law in \eqref{eq:spUpdate2}. In other words, with the periodic use of a high-level planner, the update law above can be used even in cases where the environmental disturbances are not isolated.

\section{Experiments}
\label{sec:experiments}

\begin{figure}
    \centering
    \subfloat[]{\label{subfig:exp1Init}\includegraphics[width=0.24\textwidth]{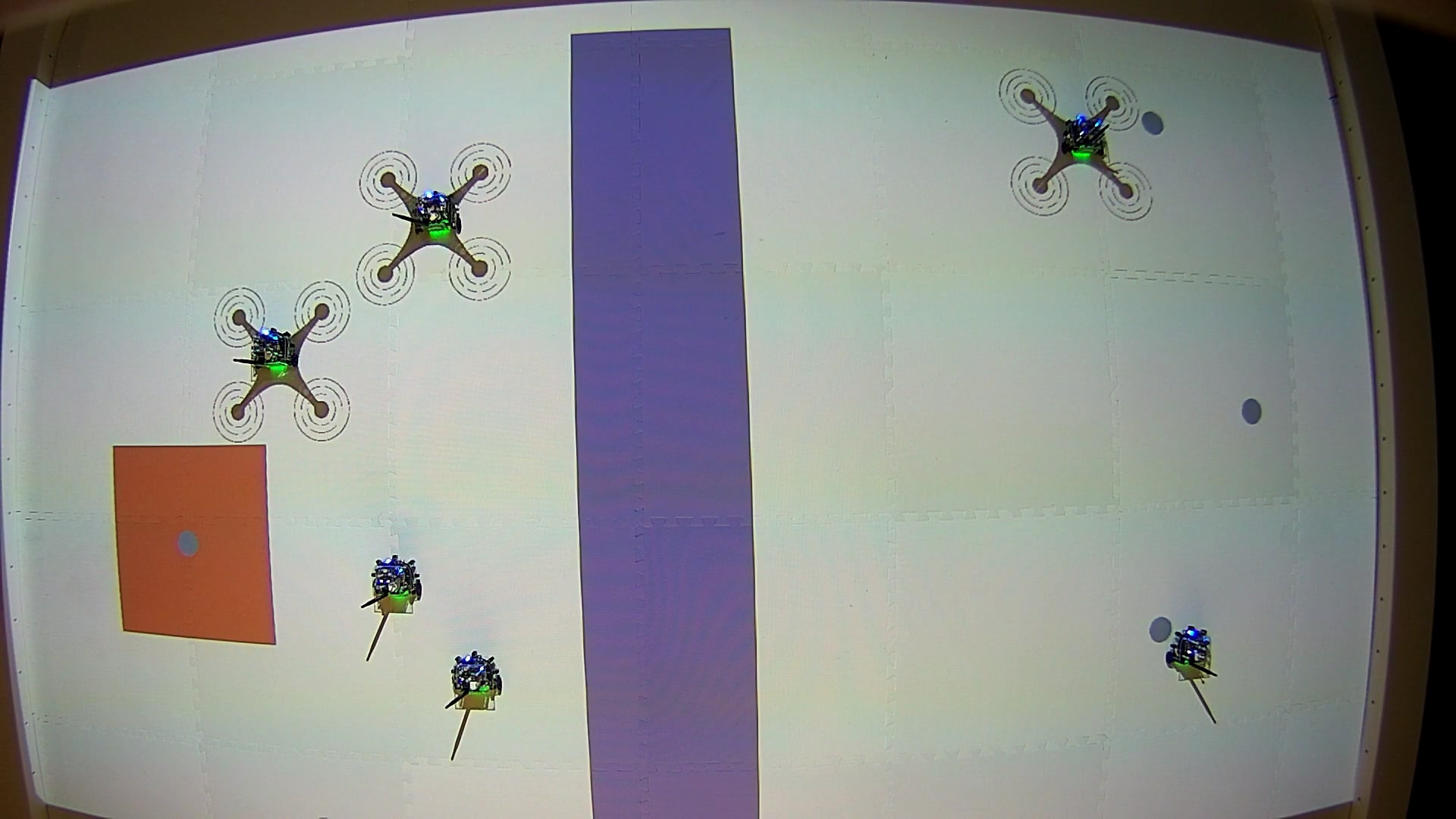}}
    \subfloat[]{\label{subfig:exp1Final}\includegraphics[width=0.24\textwidth]{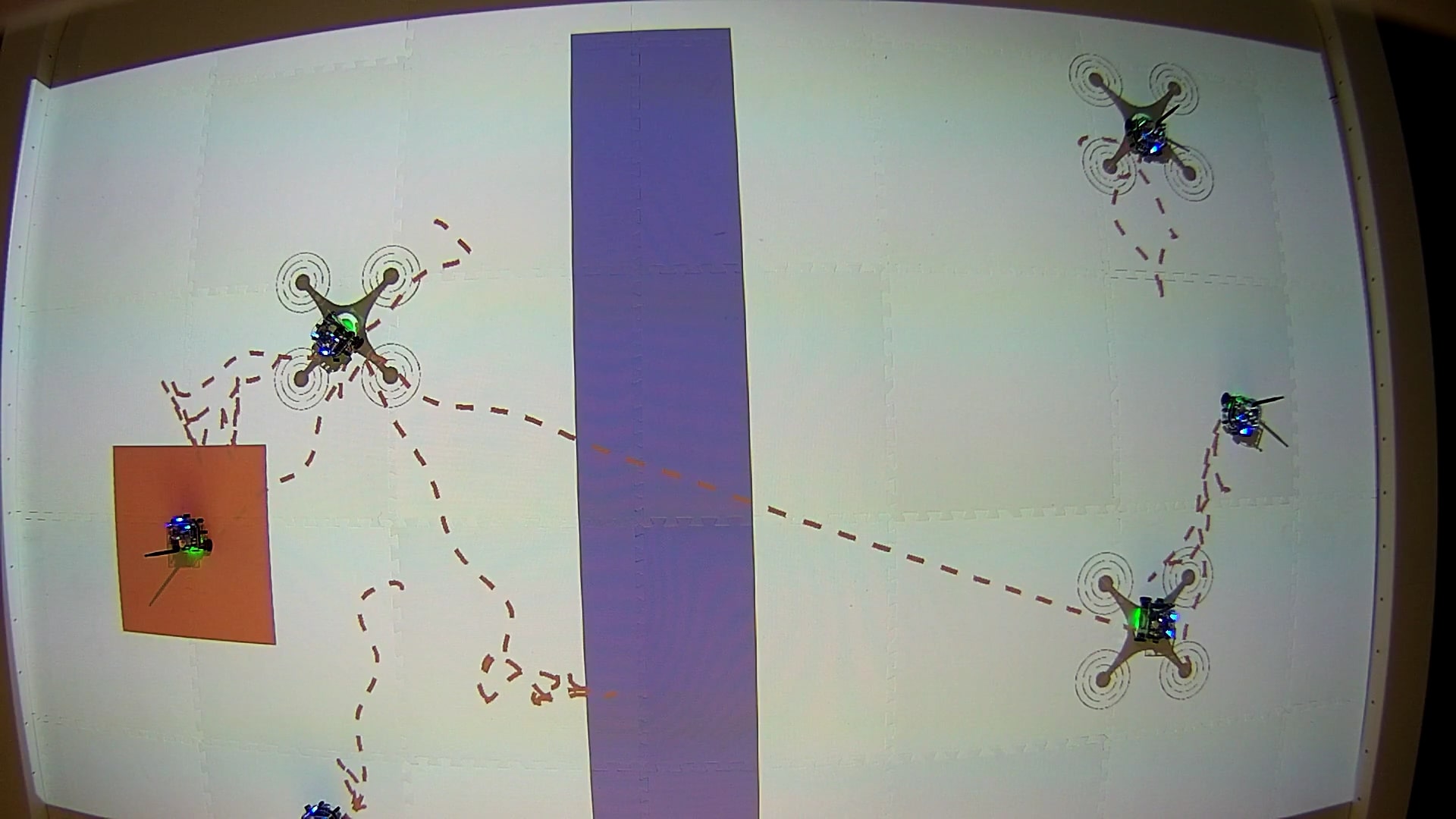}}
    \caption{A team of $6$ robots composed of $3$ quadcopter and $3$ ground robots is to execute $6$ go-to-goal tasks (blue circles). The simulated unknown environmental variations are a no-fly zone (orange rectangle) that no quadcopter can enter, and a river (blue rectangle) that no ground robot can traverse. Since the robots are unaware of these environmental features, the initial task assignment is doomed to fail. However, thanks to the update rule presented in \eqref{eq:spUpdate1}, the robots update their specialization parameters, allowing for a dynamic re-assignment of the tasks among robots. As soon by the dashed trajectories of the robots in \ref{subfig:exp1Final}, all tasks are accomplished after the re-assignments.}
    \label{fig:expA}
\end{figure}

\begin{figure}
    \centering
    \subfloat[]{\label{subfig:exp2Init}\includegraphics[width=0.16\textwidth]{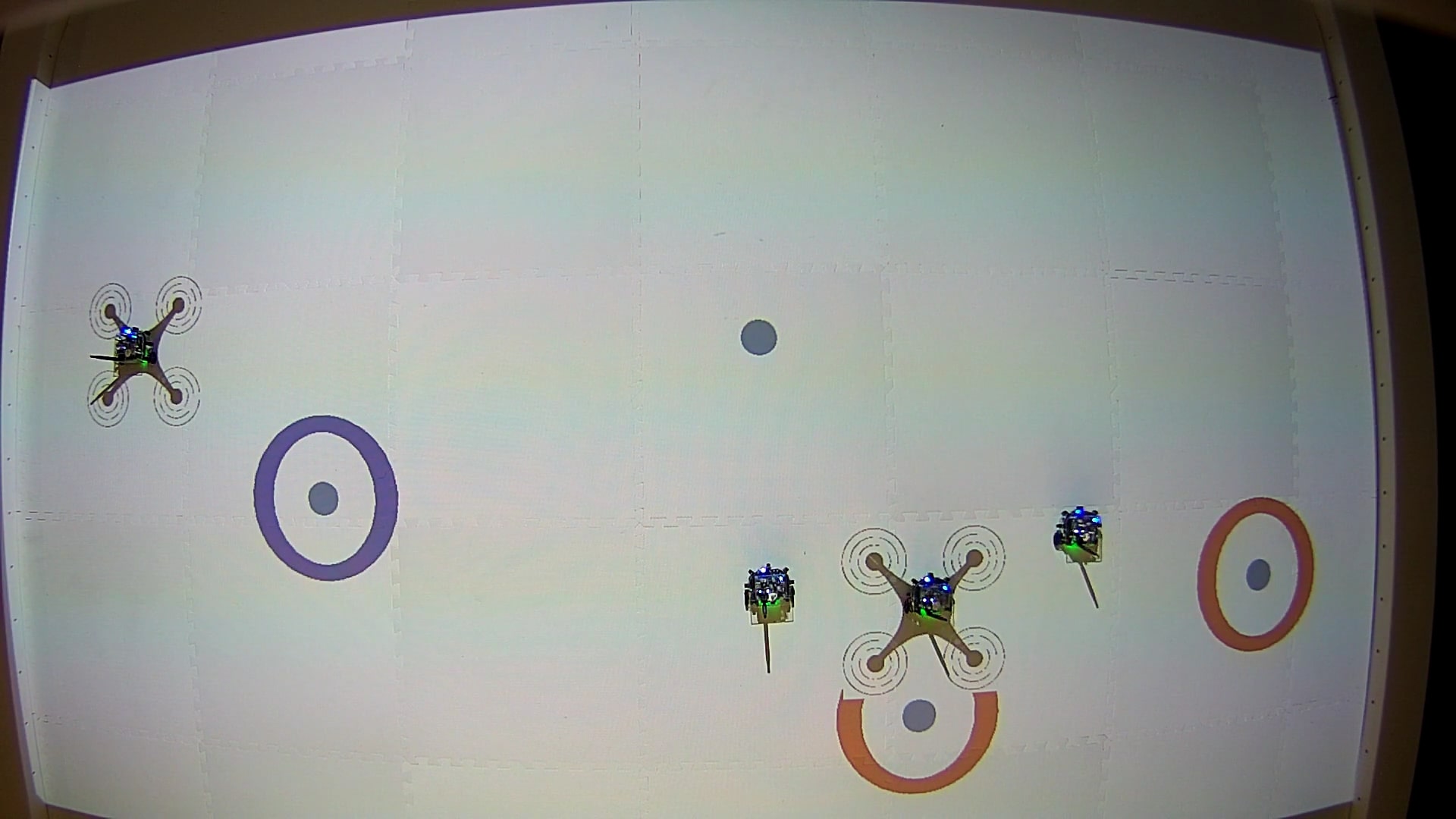}}
    \subfloat[]{\label{subfig:exp2Interim}\includegraphics[width=0.16\textwidth]{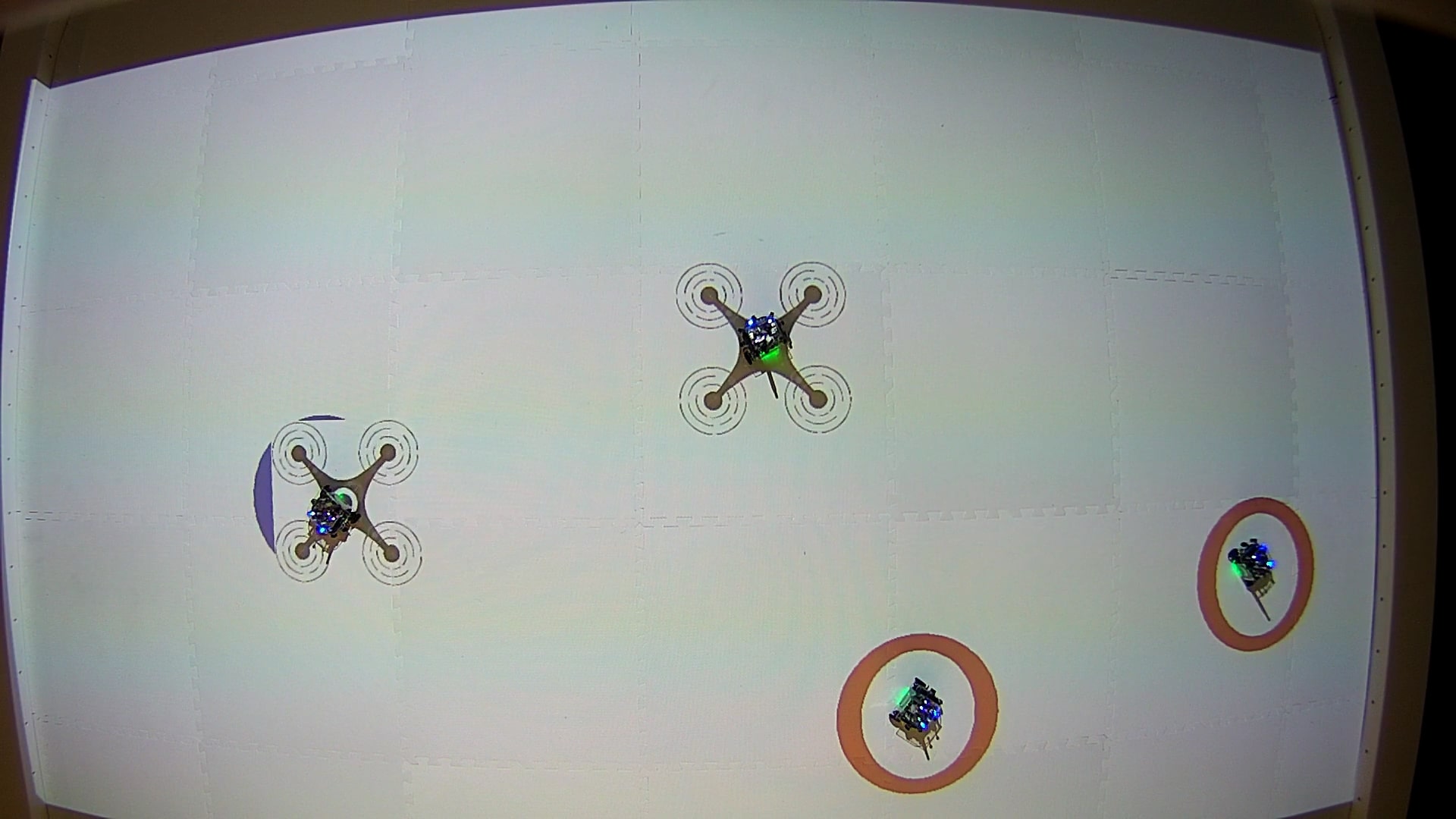}}
    \subfloat[]{\label{subfig:exp2Final}\includegraphics[width=0.16\textwidth]{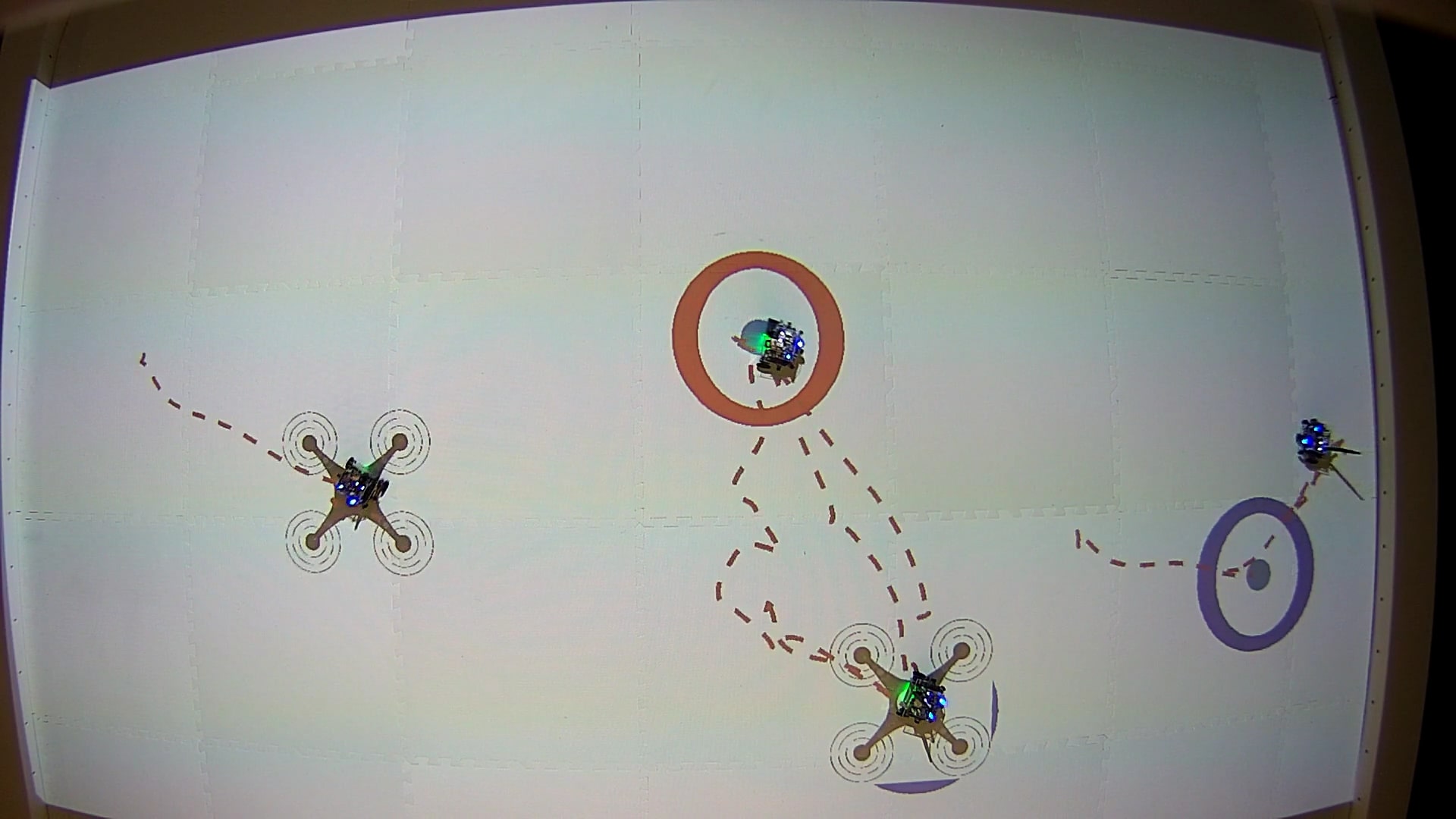}}
    \caption{A team composed of $2$ simulated quadcopters and $2$ ground robots is to execute $4$ go-to-goal tasks. Each target has a corresponding state which indicates if it is a no-fly zone (orange circle), a no-go zone for ground robots (blue circle) or if it can be accessed by any robot. The robots, randomly initialized (left), are successfully allocated so that all the tasks are accomplished (middle). Then, halfway through the experiment, all the environmental states are switched---limiting the robots abilities to execute the tasks. Updating the specialization parameters using \eqref{eq:spUpdate2}, allows the favorable dynamic reassignment of tasks. The final state of the robots accomplishing the tasks is shown on the right with the trajectories illustrated by the red-dashed lines.}
    \label{fig:expB}
\end{figure}

In this section, we validate the adaptive task allocation framework developed in this paper by deploying the algorithms on a team of real robots operating on the Robotarium \cite{pickem2017}, a remotely accessible swarm robotics testbed.
\subsection{Adaptive Update Law Experiment}
To validate the adaptive update law given in \eqref{eq:spUpdate1}, $6$ robots are tasked to execute $6$ go-to-goal tasks as shown in Figure~\ref{subfig:exp1Init}. The multi-robot system executes the dynamic task allocation algorithm, as summarized in Figure \ref{fig:feedbackLoop} while using the update law in \eqref{eq:spUpdate1}. We consider two types of robots: 3 quadcopters and 3 ground robots as shown in Figure~\ref{fig:expA}. The quadcopters (which are virtual, and are represented by ground robots in the experiment) are unable to enter the red-zone which we refer to as the no-fly zone. On the other hand, the ground robots are unable to cross the river displayed as the blue rectangle. The task positions were randomly initialized except for one, which was placed into the no-fly zone. It is important to note that the robots are unaware of the restrictions on their motion. The initial positions of the robots were chosen so that a feasible final assignment of tasks to robots was possible. For an initial specialization of $s_{ij} = 1$, Figure~\ref{subfig:exp1Final} displays the final trajectories of the agents using dashed red lines. Indeed, the team of robots reached all $6$ targets in approximately $1$ minute after re-configuring themselves various times as illustrated by the sharp changes in trajectories indicating re-assignment. For example, the final value of the specialization parameter of drone $3$, which initially attempted the go-to-goal task within the no-fly zone, is $0.8248$ for that specific go-to-goal task. This indicates that the update rule in \eqref{eq:spUpdate1} successfully captured the effectiveness of the robot at performing the task, ultimately allowing for a re-assignment of tasks.

\subsection{Isolated Disturbances Experiment}

In this subsection, we validate the update law in \eqref{eq:spUpdate2} used for cases where the environmental changes are known to be temporary or spatially-confined. As shown in Figure~\ref{subfig:exp2Init}, we consider a team composed of $2$ ground robots and $2$ simulated quadcopters, and $4$ go-to-goal tasks. Each task has an environmental state signifying whether it is a no-fly zone for the quadcopters (orange circle), a no-go zone for ground robots (blue circles) or if it is an accessible terrain for both types of robots (no circle). The environmental states of all the tasks are randomly switched halfway through the experiment to test whether the robots can re-configure themselves successfully in a dynamic environment. Shown in Figures~\ref{subfig:exp2Interim}~and~ \ref{subfig:exp2Final} are the execution of the tasks by the team of robots before and after the switch respectively. Before the switch the robots successfully execute all $4$ tasks. After the switch, the robots are reassigned to tasks and $3$ of the $4$ robots successfully accomplish their assigned task. The inability of the remaining robot to be correctly re-assigned is due to the choice of $\beta_2$ from \eqref{eq:spUpdate2}. Since the framework is agnostic to the nature of the environmental changes and therefore their rate of occurrence, a robot may be re-assigned to the same task it is unable to execute due to the effect of the integral term in the update law in \eqref{eq:spUpdate2}. \par 
In both of the above investigated cases, the dynamic task allocation algorithm reassigned the tasks among robots, so that the robots could accomplish the tasks. In both situations, the robots did not have any knowledge about the environmental variations occurring, and only evaluated the progress they were making towards accomplishing the tasks.
\section{Conclusion} \label{sec:conclusion}

This paper introduces an adaptive task allocation and execution framework for heterogeneous teams of robots which perform a set of given tasks. This was achieved by updating a parameter which encodes the effectiveness of the robots at accomplishing the tasks, thereby allowing the task allocation algorithm to reassign tasks to robots based on their current capabilities towards perfroming the different tasks. Experimental results showcase the efficacy of the approach in various multi-robot experiments conducted on the Robotarium.

\bibliographystyle{unsrt}
\bibliography{ref.bib}

\end{document}